\documentclass[10pt,twocolumn,letterpaper]{article}

\usepackage{cvpr}
\usepackage{times}
\usepackage{epsfig}
\usepackage{graphicx}
\usepackage{amsmath}
\usepackage{amssymb}
\usepackage{multirow}
\usepackage{booktabs}
\usepackage{threeparttable}

\def\ie{{\em i.e.}}
\def\eg{{\em e.g.}}

\newcommand{\bl}[1]{\textbf{#1}}
\newcommand{\mc}[1]{\mathcal{#1}}

\usepackage[pagebackref=true,breaklinks=true,letterpaper=true,colorlinks,bookmarks=false]{hyperref}

\cvprfinalcopy 

\ifcvprfinal\fi
\begin{document}

\title{Visual Attention on the Sun: What Do Existing Models Actually Predict?}

\author{Jia Li$^{1,2*}$, Daowei Li$^{1,2}$, Kui Fu$^1$ and Long Xu$^2$\\
$^1$State Key Laboratory of Virtual Reality Technology and Systems, SCSE, Beihang University\\
$^2$National Astronomical Observatories, Chinese Academy of Sciences\\
}

\maketitle

\begin{abstract}
Visual attention prediction is a classic problem that seems to be well addressed in the deep learning era. One compelling concern, however, gradually arise along with the rapidly growing performance scores over existing visual attention datasets: do existing deep models really capture the inherent mechanism of human visual attention? To address this concern, this paper proposes a new dataset, named VASUN, that records the free-viewing human attention on solar images. Different from previous datasets, images in VASUN contain many irregular visual patterns that existing deep models have never seen. By benchmarking existing models on VASUN, we find the performances of many state-of-the-art deep models drop remarkably, while many classic shallow models perform impressively. From these results, we find that the significant performance advance of existing deep attention models may come from their capabilities of memorizing and predicting the occurrence of some specific visual patterns other than learning the inherent mechanism of human visual attention. In addition, we also train several baseline models on VASUN to demonstrate the feasibility and key issues of predicting visual attention on the sun. These baseline models, together with the proposed dataset, can be used to revisit the problem of visual attention prediction from a novel perspective that are complementary to existing ones.
\end{abstract}

\let\thefootnote\relax\footnotetext{*Jia Li is the corresponding author. URL: http://cvteam.net}

\section{Introduction}
The problem of visual attention prediction has been extensively studied in the past two decades. From the early bio-inspired attention models~\cite{itti1998model,harel2007graph,zhang2008sun} to subsequent shallow learning models \cite{MIT1003,zhao2012learning,kienzle2007nonparametric} as well as the latest deep learning models~\cite{kruthiventi2017deepfix,huang2015salicon,liu2018learning,pan2016shallow,wang2018deep}, their capability advances steadily in predicting attention regions reflected by the human fixations recorded in free-viewing eye-tracking experiments. To date, the performance of the latest model on the public saliency benchmark \cite{MIT300} seems to be sufficiently high and the problem seems to be well addressed. However, a compelling concern naturally arises: what do existing deep attention models actually predict?

\begin{figure}[t]
\begin{center}
   \includegraphics[width=1.00\linewidth]{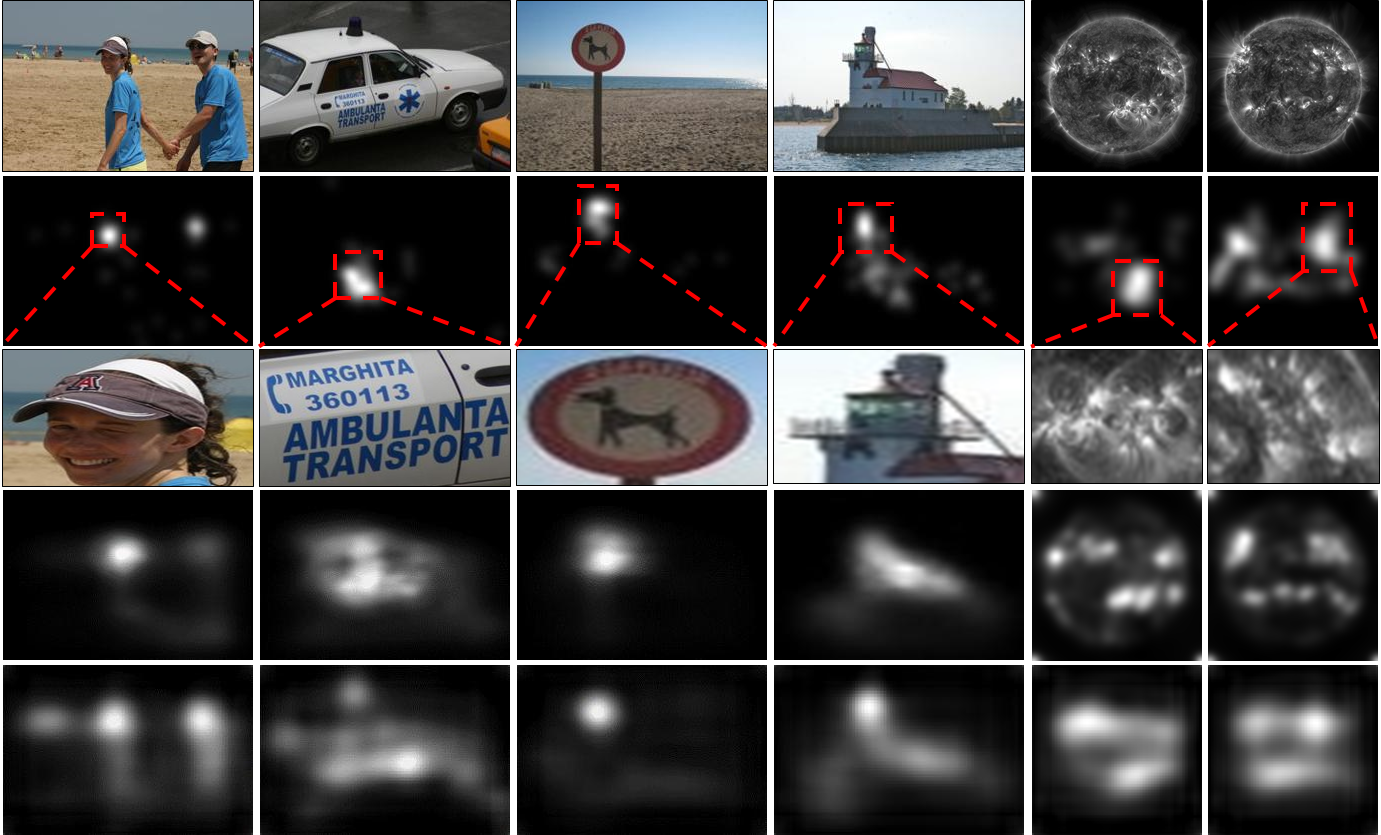}
\end{center}
   \caption{Representative results of visual attention models on ground-level and solar images. The first row shows four representative images from existing visual attention datasets and two solar images.  The second row shows the ground-truth attention maps (\ie, fixation density maps), and the third row shows the most attractive visual patterns. The last two rows show the results from the non-deep attention model SP~\cite{li2014visual} and the deep attention model SalNet~\cite{pan2016shallow}. When the non-deep model demonstrates consistent capabilities in handling ground-level and solar images, deep models fail on solar images, implying that they may only memorize specific visual patterns other than capture the inherent mechanism of human visual attention. }
\label{fig:motivation}
\end{figure}

To address this concern, we carefully inspect the images in many prevailing visual attention datasets such as MIT300~\cite{MIT300}, MIT1003~\cite{MIT1003}, CAT2000~\cite{CAT2000} and Toronto~\cite{Toronto}. We find that one common attribute shared by most images in these datasets is that they are captured on the Earth, fulfilled with repeatedly occurrence of attractive visual patterns from people, vehicle, building, etc. In most cases, the higher capability of memorizing these patterns and learning related representations, the better performance of an attention model (as shown in Fig.~\ref{fig:motivation}). However, one issue remains unclear even for the latest deep models: whether such capability sufficiently reflects the inherent mechanism of the human visual attention system?

To further investigate this issue, we propose to construct a new visual attention dataset composed of visual patterns that rarely appear in existing datasets. To maximize the diversity of visual patterns contained in the new dataset and existing ones, we refer to the solar images provided by the LSDO dataset~\cite{kucuk2017large} that are taken at specific wavelengths to record the solar activities. As shown in the last two columns of Fig.~\ref{fig:motivation}, visual patterns in these solar images are remarkably different from the classic patterns of people, vehicle, building and road signs. From the LSDO dataset, we uniformly sample 1070 images captured at the wavelength {171\AA} that perform the best in reflecting the solar active regions over three years (\ie, 2012 to 2014). For these images, we design an eye-tracking experiment to record the visual attention of 16 subjects under free-viewing conditions. From Fig.~\ref{fig:motivation}, we can see that the human-being is still capable to quickly locate the most attractive regions-of-interest even they never seen the solar images before. The question is, whether the attention models, especially the deep models that have impressive capability on ground-level images, still work well on the solar images?

To answer this question, we benchmark the performance of 17 visual attention models over the new dataset (denoted as VASUN) by using five representative evaluation metrics. From the benchmark results, we find an interesting performance reversion over the non-deep and deep models. In most of the five evaluation metrics, the performances of many state-of-the-art deep models drop remarkably. Instead, many classic shallow models that are bio-inspired or non-deep learning-based still perform impressively on solar images and significantly outperform the deep models. From these results, we find that the performance advance of existing deep attention models may mainly come from their capabilities of memorizing and predicting the occurrence of some specific visual patterns other than learning the inherent mechanism of human visual attention. In other words, most of the existing deep models, despite their impressive performances on existing ground-level datasets, still lack the capability of modeling the inherent mechanism of human visual attention. Moreover, such capability can be quantitatively measured by their performance variations on ground-level images and the solar images in VASUN. In addition, we also train several baseline models on VASUN to demonstrate the feasibility and key issues of predicting visual attention on the sun. These baseline models, together with the proposed dataset, can be used to revisit the problem of visual attention prediction from a novel perspective that are complementary to existing ones.

Our contributions are summarized as follows: 1)~We propose a novel visual attention dataset on solar images, which can be used to measure the degree that an attention model captures the inherent mechanism of human visual attention; 2)~we conduct an extensive benchmark to provide a tentative guess on what do attention models really predict; 3)~we propose five baseline models to show that visual attention on the sun is predictable and conclude several key factors that can be reused in designing new models.

\section{Related Work}
There exist many reviews about visual attention models~\cite{borji2013state} and evaluation metrics~\cite{Bylinskii2018metric}. Here we briefly review related works from two perspectives: visual attention models and benchmark datasets as well as evaluation metrics.
\subsection{Visual Attention Models}
The development of visual attention models can be roughly divided into three stages characterized as bio-inspired, shallow learning and deep learning, respectively. It is widely acknowledged that the first bio-inspired stage originates from \cite{koch1987shifts,itti1998model}. In nearly a decade after the publication of \cite{itti1998model}, a widely adopted paradigm of designing attention models is to find neurobiological evidences or hypotheses first and then mimic them with computational attention models. Such evidences or hypotheses can be represented by features such as local contrast~\cite{harel2007graph} and global irregularity~\cite{zhang2008sun}. Based on these features, a bio-inspired attention model usually acts as a feature fusion or selection strategies. Generally speaking, each of these features and mapping strategies can reflect the mechanism human visual attention from a specific perspective but may suffer from the incompleteness.

In the second stage, researchers realize that the heuristically designed mapping strategies may be insufficient to combine various attention features. Therefore, they propose to find a better feature combination strategy through shallow learning algorithms. Most of the learning algorithms adopt the supervised framework such as Support Vector Machine~\cite{MIT1003} and Adaptive Boosting~\cite{zhao2012learning}. In addition, some algorithms adopt unsupervised frameworks such as sparse coding~\cite{hou2009dynamic} and statistical learning~\cite{li2014visual}. In particular, many datasets are proposed along with models in the second stage to meet the demand of machine learning algorithms.

With the cumulation of various visual attention datasets, models propose to direct learn a deep network that maps the input image to the output attention map. Compared with bio-inspired and shallow learning models, deep models such as eDN~\cite{eDN}, DeepGaze~\cite{kummerer2014deep}, DeepFix~\cite{kruthiventi2017deepfix}, SALICON~\cite{huang2015salicon}, SalNet~\cite{pan2016shallow}, DVA~\cite{wang2018deep} and SAM-ResNet\cite{SAMresnet} achieved much better results by learning better feature representations via novel deep learning architectures~\cite{wang2018deep}, multi-scale input strategy~\cite{li2015visual} and specially designed loss functions~\cite{jetley2016end}. From a positive perspective of view, these deep models do achieve impressive performance scores in existing benchmarks that can be hardly reached by bio-inspired and shallow learning models. From a negative perspective of view, however, it is not clear whether these deep models really capture inherent mechanism of human visual attention. Although some works have explore the relationships between deep networks with neural responses~\cite{yamins2014performance,khaligh2014deep}, a compelling concern still arises in the literature: what do existing deep attention models actually predict?


\begin{table}[t]
\caption{The subject number and image resolution of representative image datasets and our VASUN dataset. \#Img and \#Sub mean the number of images and subjects in the dataset.}
\centering{
\begin{tabular}{cccc}
\toprule
Dataset   & \#Img & \#Sub. & Max Res.\\
\midrule
MIT300~\cite{MIT300}      & 300     & 39  & {1024$\times{}$1024}\\
MIT1003~\cite{MIT1003}    & 1003    & 15  & {1024$\times{}$1024}\\
Toronto~\cite{Toronto}    & 120     & 20  & 511$\times$681\\
CAT2000~\cite{CAT2000}    & 2000    & 18  & 1920$\times{}$1080\\
SALICON~\cite{SALICON}    &  15000  & -   & 640$\times{}$480\\
PASCAL-S~\cite{PASCAL-S}  & 850     & 8   & {500$\times{}$500}\\
DUT-O~\cite{DUT-O}        & 5168    & 5   & {401$\times{}$401}\\
\hline
VASUN                     & 1070    & 16  & 1024$\times{}$1024\\
\bottomrule
\end{tabular}
}
\label{tab:dataset}
\end{table}
\subsection{Benchmark Datasets and Metrics}
The evaluation of visual attention models is also a challenging task since the consistency between human fixations and model predictions can be measured from many perspectives~\cite{Bylinskii2018metric}. As a result, many benchmark datasets have been proposed in the literature. Among these datasets, some of them are specifically designed for studying the human fixation prediction problem of more than a dozen subjects in the free-viewing conditions~(\eg, MIT300~\cite{MIT300}, MIT1003~\cite{MIT1003}, Toronto~\cite{Toronto} and CAT2000~\cite{CAT2000}), while the rest ones just record fixations of several subjects to facilitate subsequent annotations (\eg, Pascal-S~\cite{PASCAL-S} and DUT-O~\cite{DUT-O}). The number of subjects and their image resolutions of these datasets can be found in Table~\ref{tab:dataset}.

Given a dataset, there are many ways to evaluate an attention model, leading to a dozen of evaluation metrics~\cite{riche2013saliency,li2015metric}. Here we only introduce the most representative five metrics that will be used in this study.

1)~Area Under the ROC Curve (AUC) and Shuffled AUC (sAUC). In computing AUC, the recorded fixations are selected as positives, while all the other locations are treated as negatives. After that, the true positive rates and false positive rates are computed at all probable thresholds to plot a ROC curve to compute the AUC score. The main difference between sAUC and AUC is that the negatives correspond to fixations shuffled from other images in the same dataset, leading to a more strict evaluation standard.

2)~Normalized Scanpath Saliency (NSS). NSS is considered to be one of the most effective metrics in many recent studies~\cite{riche2013saliency,Bylinskii2018metric}. It is computed by first normalizing the predicted attention map to zero mean and unit standard deviation, and then computes the average responses at fixations.

3)~Similarity (SIM) and Correlation Coefficient (CC). The ground-truth and predicted attention maps can be also treated as probability distributions. As a result, SIM can be computed as their intersection and CC describes their linear relationship.

\section{The VASUN Dataset}
In previous benchmarks of attention models
\cite{borji2013state}, the generalization ability of models are less discussed since almost all datasets are fulfilled with the same categories of scenarios (\eg, people, vehicles, buildings and road signs). To measure the generalization ability, we need to construct a dataset that contains new visual patterns. Toward this end, we refer to the LSDO dataset~\cite{kucuk2017large}, which is a large scale dataset that contains 260K solar images captured in three years by the National Aeronautics Space Agency (NASA) in the Solar Dynamics Observatory (SDO) mission. This dataset consists of $4096\times{}4096$ solar images captured at three wavelengths (\ie, 94 {\AA}, 171 {\AA} and 193 {\AA}). Neither such solar images nor their visual patterns have appear in any existing image attention datasets.

From the LSDO dataset, we construct a new image attention dataset to study visual attention on the sun (denoted as VASUN). Since the main objective of this dataset is to measure the generalization ability of attention models, we adopt the similar setting with MIT1003, the most widely used dataset with ground-level scenarios. That is, we sample 1070 images taken at the wavelength 171\AA ~from LSDO and down-sample them to the resolution of 1024$\times$1024. On these images, we record the visual attention of 16 subjects in eye-tracking experiments. These 16 subjects (10 males and 6 females) have normal or correct-to-normal visions. None of them has prior knowledge of astronomy and solar physics to avoid subjective bias.

In the experiment, each subject is asked to free-view every image for 4 seconds on a 22-inch color monitor. A chin rest is utilized to reduce head movements and enforce a viewing distance of 75 cm. During the free-viewing process, an eye-tracking apparatus with a sample rate of 500 Hz (SMI RED 500) is used to record various types of eye movements. Finally, we keep only the fixations and generate the ground-truth attention map $G$ for an image $I$ by overlaying Gaussian blobs centered at these fixations:
\begin{equation}
G(x,y)=\sum_{f=1}^{N}\frac{1}{2\pi{}\sigma^2}\exp\left(-\frac{(x-x_f)^2+(y-y_f)^2}{2\sigma^2}\right),
\end{equation}
where $\{(x_f,y_f),f=1,\ldots,N\}$ is the recorded fixations (each lasts about two milliseconds). $\sigma$ is a parameter computed according to the viewing distance and angle. Note that these ground-truth attention maps are then normalized into the dynamic range of $[0,1]$.

\begin{figure}[t]
\begin{center}
   \includegraphics[width=1.00\linewidth]{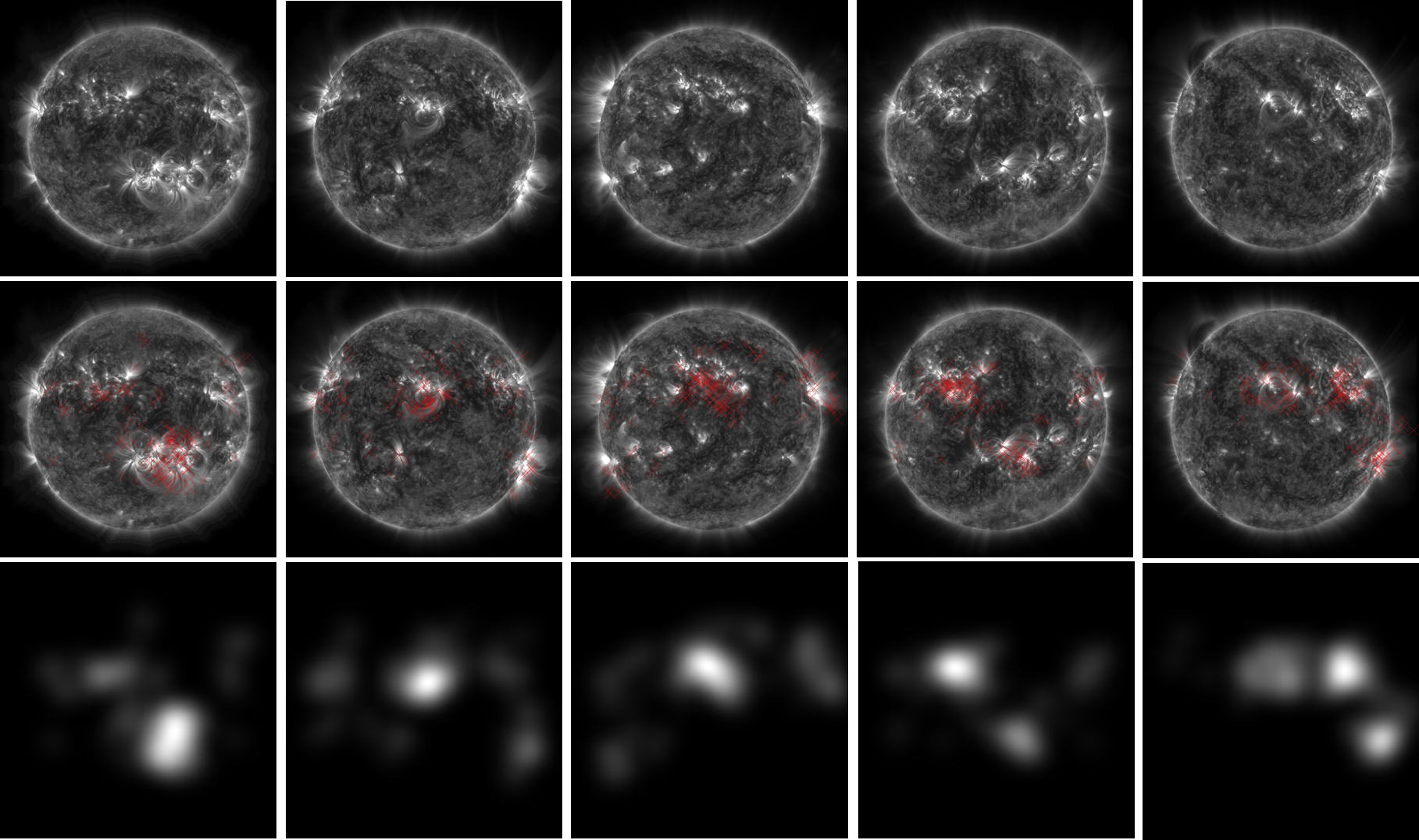}
\end{center}
   \caption{Representative examples from VASUN. Red dots in the second row indicate the recorded fixation points, which are used to generate the ground-truth attention maps in the third row.}
\label{fig:datasetexample}
\end{figure}

\begin{figure}[t]
\begin{center}
   \includegraphics[width=1.00\linewidth]{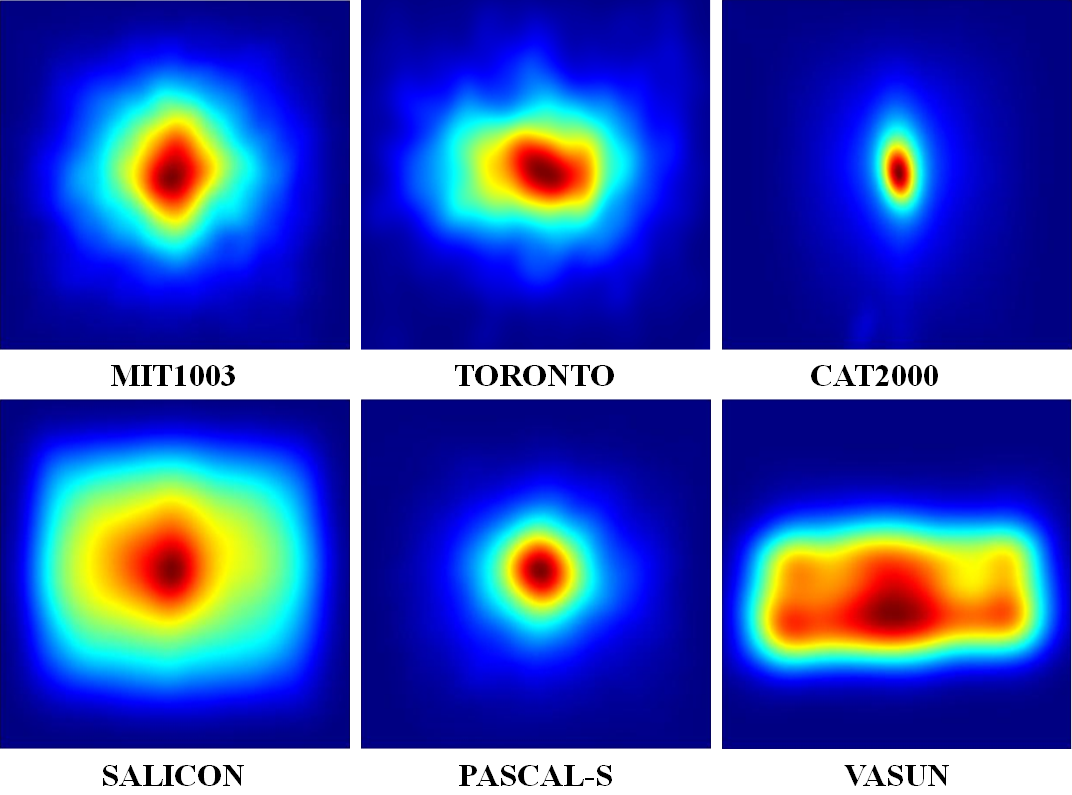}
\end{center}
   \caption{The average attention maps of various datasets. We can see that the distribution of human visual attention on VASUN is quite different from those in previous datasets, making a good dataset to test model generalization ability. }
\label{fig:aam}
\end{figure}

Some representative examples of the recorded fixations and ground-truth attention maps can be found in Fig.~\ref{fig:datasetexample}. From Fig.~\ref{fig:datasetexample}, we can see that most human visual attention is allocated to large active regions, making the visual attention prediction task on solar images a theoretically simple task. In addition, we also compare the average attention maps of VASUN with previous datasets in Fig.~\ref{fig:aam}. We can see that the distribution of fixations in VASUN is quite different from previous ground-level datasets. This may be caused by the fact that the solar images have no photographer bias that tend to place the target at image centers. This phenomena further validates that VASUN can be used to test the generalization ability of visual attention models.

\section{Model Benchmark}
To test the generalization ability of existing visual attention models and find what they actually predict, we propose to measure their performance variations over VASUN and MIT1003. Note that MIT1003 contains 1003 images annotated by 15 subjects, which is quite similar to VASUN.

\begin{figure*}[t]
\begin{center}
\includegraphics[width=1.00\textwidth]{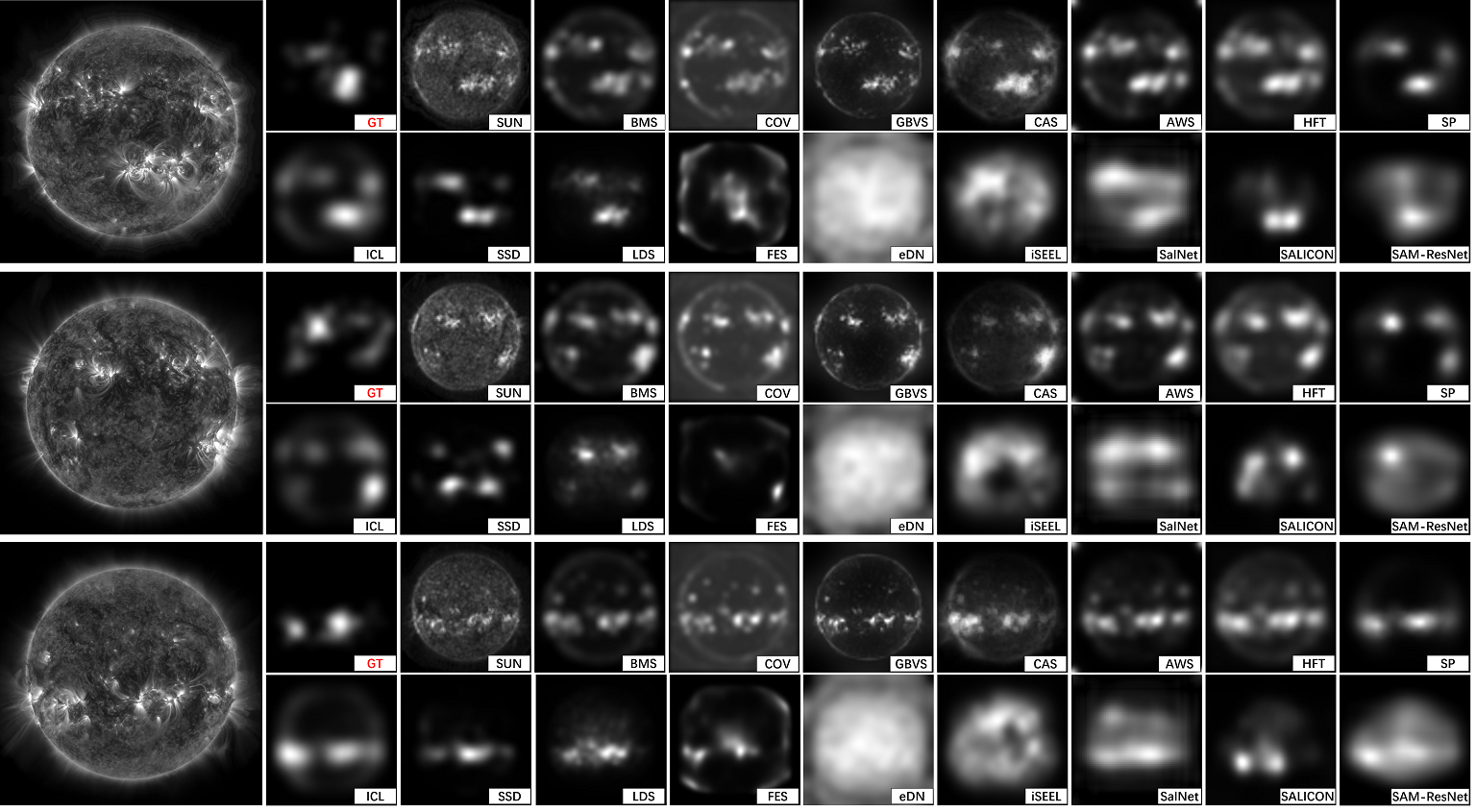}
\end{center}
\caption{Representative results of 17 attention models on VASUN. GT means ground-truth.}
\label{fig:benchmarkResult}
\end{figure*}

On these two datasets, we benchmark 17 visual attention models, which can be roughly divided into three groups: 1)~the \bl{BIO} group contains seven bio-inspired models, including SUN~\cite{zhang2008sun}, BMS~\cite{zhang2013saliency}, COV~\cite{erdem2013visual}, GBVS~\cite{harel2007graph}, CAS~\cite{CAS}, HFT~\cite{HFT} and AWS~\cite{Garciadiaz2012On}; 2)~the \bl{SL} group contains five shallow learning models, including ICL~\cite{hou2009dynamic}, SP~\cite{li2014visual}, SSD~\cite{SSD}, LDS~\cite{LDS} and FES~\cite{FES}; 3)~the \bl{DL} group contains five deep learning models, including  eDN~\cite{eDN}, iSEEL~\cite{R2016Exploiting}, SalNet~\cite{pan2016shallow}, SALICON~\cite{huang2015salicon} and SAM-ResNet~\cite{SAMresnet}. All these models have public source code on the Internet, and we use their default parameters to generate the attention maps. The predictions of these models are evaluated using five metrics, including AUC, sAUC, NSS, SIM and CC. The performance scores can be found in Table~\ref{tab:benchmarkScore}. Some representative results on VASUN can be found in Fig.~\ref{fig:benchmarkResult}.

From Table~\ref{tab:benchmarkScore}, we compare the performances of models from different groups. We find that on the MIT1003 dataset, deep attention models significantly outperform shallow learning models and bio-inspired models. However, on the VASUN dataset, we obtain a nearly reverse model ranking since all deep models suffer remarkable performance drops in terms of NSS. These results indicate that the significant performances of existing deep attention models may come from their capabilities in memorizing and detecting some specific visual patterns other than learning the inherent mechanism of human visual attention.

To further validate this point, we investigate the models in the three groups. For the bio-inspired group, the best-performing model over MIT1003 is GBVS, while the best model over VASUN changes to HFT. Note that GBVS is also the only bio-inspired model that suffers a performance drop when being applied on VASUN. This may be caused by the fact that the attention regions on solar images may have remarkably different sizes. As a result, the random walk of GBVS over graph nodes formed by fix-sized blocks may be insufficient to depict over-large and over-small attention regions. By contrast, HFT conducts scale-space analysis to select the most propriae scales, which enhances its capability to handle attention regions with various sizes.

\begin{table*}[t]
\caption{Benchmark of 17 models with default parameters. Bold means the highest score in each group.}
\centering{
\begin{tabular}{p{0.1cm}ccccccp{0.005cm}cccccc}
\toprule
\multicolumn{2}{c}{\multirow{2}*{Models}} & \multicolumn{5}{c}{MIT1003} && \multicolumn{5}{c}{VASUN}\\
\cline{3-7} \cline{9-13}
    & & AUC & sAUC & NSS & SIM & CC && AUC & sAUC & NSS & SIM & CC \\
\midrule
  \bl{\multirow{7}{*}{\rotatebox{90}{BIO}}}
    & SUN~\cite{zhang2008sun}      & 0.655 & 0.591 & 0.668 & 0.269 & 0.212 && 0.844 & 0.664 & 1.135 & 0.443 & 0.488\\
    & BMS~\cite{zhang2013saliency} & 0.790 & \bl{0.690} & 1.250 & 0.330 & 0.360 && 0.838 & 0.717 & 1.333 & 0.500 & 0.569\\
    & COV~\cite{erdem2013visual}   & 0.752 & 0.619 & 0.899 & 0.258 & 0.285 && 0.792 & 0.679 & 1.139 & 0.378 & 0.487\\
    & GBVS~\cite{harel2007graph}   & \bl{0.830} & 0.660 &\bl{1.380} & 0.360  & \bl{0.420} && 0.821 & 0.673 & 1.146 & 0.431 & 0.489\\
    & CAS~\cite{CAS}               & 0.758 & 0.632 & 0.992 & 0.320 & 0.310 && \bl{0.893} & 0.722 & 1.372 & 0.515 & 0.584\\
    & AWS~\cite{Garciadiaz2012On}  & 0.730 & 0.652 & 1.039 & 0.316 & 0.321 && 0.871 & 0.743 & 1.474 & \bl{0.537} & 0.629\\
    & HFT~\cite{HFT}               & 0.828 & 0.663 & 1.326 & {\bl{0.375}} & 0.419 && 0.887 & {\bl{0.749}} & {\bl{1.528}} & 0.522 & \bl{0.653}\\
\hline
  \bl{\multirow{5}{*}{\rotatebox{90}{SL}}}
    & ICL~\cite{hou2009dynamic}    & 0.605 & 0.515 & 0.422 & 0.244 & 0.130 && 0.874 & 0.767 & 1.564 & 0.556 & 0.665\\
    & SP~\cite{li2014visual}       & 0.827 & 0.663 & 1.372 & 0.401 & 0.434 && {\bl{0.893}} & {\bl{0.787}} & {\bl{1.768}} & {\bl{0.626}} & \bl{0.749}\\
    & SSD~\cite{SSD} & 0.787       & 0.651 & 1.342 & 0.398 & 0.427 && 0.858 & 0.717 & 1.501 & 0.560 & 0.633\\
    & LDS~\cite{LDS}               & {\bl{0.857}} & {\bl{0.680}} & {\bl{1.495}} & {\bl{0.434}} & {\bl{0.478}} && 0.860 & 0.734 & 1.553 & 0.571 & 0.656\\
    & FES~\cite{FES}               & 0.832 & 0.663 & 1.396 & 0.423 & 0.447 && 0.759 & 0.658 & 1.066 & 0.447 & 0.448\\
\hline
  \bl{\multirow{5}{*}{\rotatebox{90}{DL}}}
    & eDN~\cite{eDN}               & 0.851 & 0.657 & 1.292 & 0.304 & 0.408 && 0.837 & 0.698 & 1.078 & 0.378 & 0.464\\
    & iSEEL~\cite{R2016Exploiting} & 0.876 & \bl{0.743} & 1.870 & 0.460 & 0.587  && 0.841 & 0.648 & 0.987 & 0.451 & 0.425\\
    & SalNet~\cite{pan2016shallow} & 0.863 & 0.724 & 1.619 & 0.395 & 0.512 && 0.877 & 0.729 & 1.361 & 0.512 & 0.580\\
    & SALICON~\cite{huang2015salicon} & 0.870 & 0.740 & 2.120 & 0.600 & 0.740 && 0.857 & 0.705 & 1.400 & {\bl{0.537}} & 0.592\\
    & SAM-ResNet~\cite{SAMresnet}      & \bl{0.913} & 0.617 & {\bl{2.893}} & \bl{0.708} & \bl{0.768} && \bl{0.894} & \bl{0.743} & \bl{1.408} & 0.533 & \bl{0.599}\\
\bottomrule
\end{tabular}
}
\label{tab:benchmarkScore}
\end{table*}

Different from the bio-inspired models, shallow learning methods try to detect attention regions by learning the prior knowledge or fusing various pre-attentive features by machine learning methods based on thousands or even millions of images. As a result, models in the shallow-learning group generally perform better than the bio-inspired models on both datasets.  In addition, the best shallow learning model on MIT1003 is LDS, while SP becomes the best one over VASUN. This may be caused by the way of learning prior knowledge. LDS is a classic learning method that constructs a discriminative subspace on a small amount of images, while SP learns the foreground prior and the correlation prior from millions of images by modeling the spatial distributions and correlations of various visual stimuli. In this manner, the prior knowledge learned by SP becomes more robust than LDS and thus can be applied on various scenarios. In particular, SP outperforms all the other 16 models from the three groups, which further validates that the prior knowledge from big data is more robust and can improve the generalization ability of a model.

\begin{figure*}[t]
\begin{center}
   \includegraphics[width=1.00\textwidth]{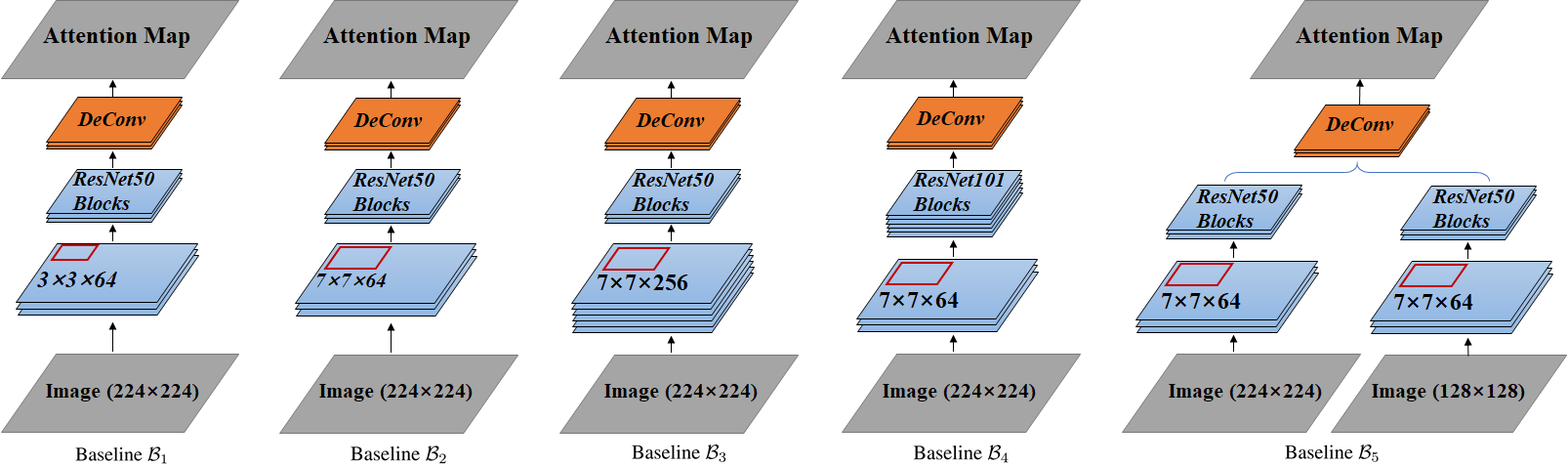}
\end{center}
   \caption{The network architectures of five baseline models.}
\label{fig:baselines}
\end{figure*}

In the deep learning group, we find that many models suffer from a severe performance drop. In this group, the best performing model over MIT1003 is SAM-ResNet, whose performance drops remarkably on VASUN. This may be caused by the fact that the features in these deep models are learned other than being designed. While traditional features such as center-surround contrasts enhance the generalization ability of bio-inspired and shallow learning models, the deep features may over-fit to specific visual patterns that frequently appear in existing ground-level datasets (\eg, faces, vehicles and texts). Such features can demonstrate impressive generalization ability over existing ground-level datasets since the training and testing data have similar distributions of such visual patterns. However, visual patterns in ground-level and solar images have remarkably different distributions. In this case, these features are likely to fail, leading to the poor performances of deep models on solar images.

To sum up, bio-inspired and shallow learning models have better generalization ability than deep models, and a better generalization ability may come from two aspects: 1)~bio-inspired pre-attentive features, 2)~robust prior knowledge for feature fusion. In particular, the prior knowledge unsupervisedly learned from millions of images demonstrates higher generalization ability. The success of deep models over ground-level datasets may mainly comes from their capability of learning feature representations and memorizing specific visual patterns, which is still far from the inherent mechanism of human visual attention. In other words, there is still a long way to go in the area of visual attention prediction even when the performance over existing ground-level datasets seems to be saturated in recent works.

\section{In-depth Analysis of Baseline Models}
In this section, we further provide an in-depth analysis of five baseline models to reveal two things: 1)~is the visual attention on the sun really predictable? and 2)~what key factors should be taken into account in designing a deep model for predicting visual attention on the sun.

The architecture of the five baseline models used in this study are shown in Fig.~\ref{fig:baselines}, from which we can see that each model slightly differs from the previous one to reveal the influence of a specific setting. All these baseline models adopt a sub-network $\mc{B}_{ini}$ with several convolution  layers to extract low-level features first, followed by the ResNet sub-network $\mc{B}_{res}$ to extract high-level features and de-convolution layers to derive attention maps. As shown in Fig.~\ref{fig:baselines}, the first baseline model $\mc{B}_1$ adopts 64 kernels with the size $3\times{}3$ in sub-network $\mc{B}_{ini}$ and 50 layers in sub-network $\mc{B}_{res}$, while the second and third baseline models $\mc{B}_2$ and $\mc{B}_3$ change the kernels to the size $7\times{}7$ and the kernel number to 256 in the sub-network $\mc{B}_{ini}$, respectively. The forth model $\mc{B}_4$ replace the sub-network $\mc{B}_{res}$ of $\mc{B}_1$ to 101 layers, while the last model $\mc{B}_5$ further takes two input streams with different resolutions. By measuring the performance variations of different baseline models, we can gradually find the key issues that should be considered in predicting visual attention on the sun.


\begin{table*}[t]
\caption{Performance of five baseline models on VASUN-testing before and after being fine-tuned on VASUN-training. Var (\%) is the NSS variation between VASUN and MIT1003.}
\centering{
\begin{tabular}{ccccccp{0.005cm}cccccc}
\toprule
{\multirow{2}*{Models}} & \multicolumn{5}{c}{Before Fine-tuning} && \multicolumn{5}{c}{After Fine-tuning} & \multirow{2}*{Var (\%)}\\
\cline{2-6} \cline{8-12}
    & AUC & sAUC & NSS & SIM & CC && AUC & sAUC & NSS & SIM & CC \\
\midrule
    Baseline $\mc{B}_1$ & 0.827 & 0.652 & 1.001 & 0.456 & 0.431 && 0.899 & 0.797 & 1.802 & 0.655 & 0.766 & +80.02\\
    Baseline $\mc{B}_2$ & 0.820 & 0.682 & 1.158 & 0.479 & 0.493 && 0.900 & 0.802 & 1.824 & 0.660 & 0.775 & +57.51\\
    Baseline $\mc{B}_3$ & 0.847 & 0.684 & 1.157 & 0.484 & 0.495 && 0.903 & 0.796 & 1.788 & 0.648 & 0.759 & +54.54\\
    Baseline $\mc{B}_4$ & 0.847 & 0.684 & 1.157 & 0.484 & 0.495 && 0.903 & 0.796 & 1.788 & 0.648 & 0.759 & +54.54\\
    Baseline $\mc{B}_5$ &\bl{0.866} &\bl{0.718} &\bl{1.275} &\bl{0.506} &\bl{0.515} &&\bl{0.958} &\bl{0.879} &\bl{2.223}&\bl{0.852}&\bl{0.947}& +74.35\\
\bottomrule
\end{tabular}
}
\label{tab:baselineScore}
\end{table*}

\begin{figure*}[t]
\begin{center}
\includegraphics[width=1.00\textwidth]{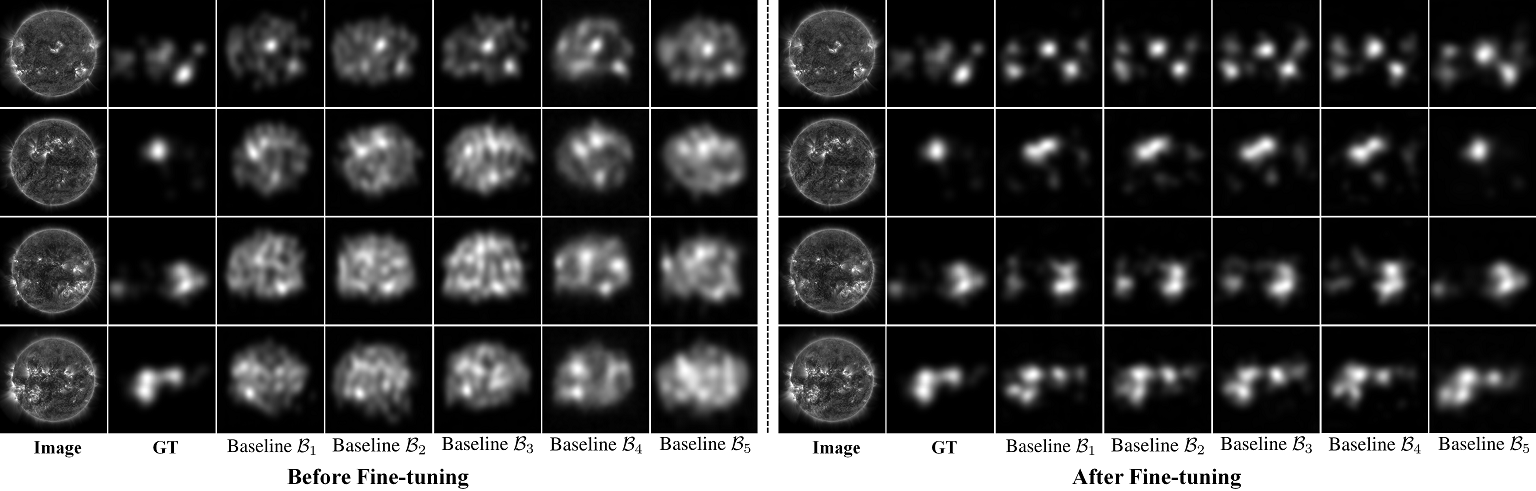}
\end{center}
\caption{Representative results of five baseline models before and after being fine-tuned on VASUN. GT means ground-truth.}
\label{fig:baselineResult}
\end{figure*}

In training the baseline model, we divide VASUN into a training subset (670 images), a validation subset (100 images) and a testing set (300 images). For the five baseline models, we first train them on the 10000 ground-level images from the SALICON dataset and then fine-tune them on VASUN. The performance scores of these five models on VASUN-testing are shown in Table~\ref{tab:baselineScore}, respectively. Some representative results of $\hat{\mc{B}}_i$ and $\mc{B}_i$ are simultaneously shown in Fig.~\ref{fig:baselineResult} to provide an intuitive comparison of what are actually predicted before and after the fine-tuning process on VASUN-training.

From Table~\ref{tab:baselineScore}, we find that all the five baseline models achieve a high performance on VASUN-testing after fine-tuning. As shown in Fig.~\ref{fig:baselineResult}, the predictions of all these five baseline models are very poor before being fine-tuned on VASUN-training. However, their AUC scores in these cases still reach up to 0.80-0.87, implying that the traditional AUC is not suitable for the performance evaluation when predicting visual attention on the sun. In particular, the baseline $\mc{B}_5$ that uses images at two-scales as the input can reach up to 0.96 in AUC and 0.95 in CC, implying perfect predictions in most cases. From these results, we can safely conclude that the visual attention on the sun is predictable, which also means the visual patterns, although quite different from those in ground-level images, are still learnable even with a simple deep network.

In addition, we conduct an analysis of these five models to show the key factors that should be considered in designing a deep model for predicting visual attention on the sun. By comparing the performance of $\mc{B}_1$ and $\mc{B}_2$, we find that $\mc{B}_2$ outperforms $\mc{B}_1$ before fine-tuning but the performances become comparable after fine-tuning. Considering that the only difference between $\mc{B}_1$ and $\mc{B}_2$ is the kernel size of the first convolution layer changes from $3\times3$ to $7\times7$. On solar images, a larger filter means that the low-level features take the contrast of bigger area into consideration, which can provide richer information for the high-level convolutional filters. Therefore, large-sized kernels should be considered with high priority in the problem of visual attention prediction on the sun.

The second comparisons are conducted between $\mc{B}_2$ and $\mc{B}_3$, which differ only in the number of kernels in the first convolution layer. Although the number of kernels in the first layer increases form 64 to 256, there is no obvious improvement in their performances on solar images. The reason is that the solar image in our dataset only has one channel, which means that 64 filters have sufficient capacity to represent the low-level features and the rest filters used in the first layer of $\mc{B}_3$ only have slight contribution during the training process. This means the problem of visual attention prediction is a simple problem with a limited number of attentive patterns. In addition, we find that $\mc{B}_2$ and $\mc{B}_3$ have the lowest performance variance before and after being fine-tuned on VASUN-training. This further validates that ground-level and solar images have different data distributions (\ie, visual patterns) so that the generalization ability favors simple models and becomes inversely proportional to the model complexity.

The baseline $\mc{B}_4$ is the reflection of the influence caused by the depth of the network. Previous work shows that the semantic information will become more richer with the increase of the network depth and deeper models usually can get a better performance. However the results in Table~\ref{tab:baselineScore} suggest that when predicting visual attention on solar images, the deeper model $\mc{B}_4$ shows a slightly lower performance than $\mc{B}_2$ that only has a half number of layers. By analyzing the features of solar images, we find that the deeper parts in a network usually lose the information contained in the low-level feature maps, while such low-level features may be sufficient to the prediction of visual attention under free-viewing conditions.

The last factor that may influence the performance of the fixation prediction is handling multi-scale input images in $\mc{B}_5$. The results show that the performance of $\mc{B}_5$ has significant improvement compared with the base model $\mc{B}_2$. As shown in Fig.~\ref{fig:baselineResult}, we can see that the predictions of $\mc{B}_5$ are often cleaner than the other four baselines. This may be caused by two factors. First, a two-stream architecture can enhance the most attractive regions twice, leading to a cleaner map. Second, the multi-scale inputs have a better capability in capturing attractive regions with various sizes.

To sum up, visual attention on the sun is predictable and prefers large convolutional kernels, simple and shallow networks as well as multi-scale inputs in model design. The types of visual patterns and the data distribution in solar images are remarkably different from those in the ground-level images, making the VASUN a good benchmark dataset for generalization ability test.

%

\section{Conclusions}
In this paper, we revisit the problem of visual attention prediction from a novel perspective: the generalization ability. Instead of testing models across various ground-level datasets, we propose a new dataset of solar images, which contains many visual patterns that rarely appear in existing datasets. Be measuring the performance variations of massive visual attention models on ground-level and solar datasets, we find that classic attention models that adopt bio-inspired or shallow learning framework show stable performances, while the deep learning models suffer a remarkable performance drop. These results imply that the deep models only memorize visual patterns that frequently appear in existing datasets, and they are still far from capturing the inherent mechanism of human visual attention. In addition, we design five baseline models to step-wisely depict the key factors in designing visual attention models on the sun. From the results of these models, we find that visual attention on the sun is predictable and conclude key factors may be necessary to improve the performance and generalization ability in attention prediction on solar image.

{\small
\bibliographystyle{ieee}
\bibliography{egbib}

\begin{thebibliography}{10}\itemsep=-1pt

\bibitem{borji2013state}
A.~Borji and L.~Itti.
\newblock State-of-the-art in visual attention modeling.
\newblock {\em IEEE TPAMI}, 35(1):185--207, 2013.

\bibitem{CAT2000}
A.~Borji and L.~Itti.
\newblock Cat2000: A large scale fixation dataset for boosting saliency
  research.
\newblock {\em CVPR 2015 workshop on "Future of Datasets"}, 2015.

\bibitem{Toronto}
N.~D.~B. Bruce and J.~K. Tsotsos.
\newblock Saliency based on information maximization.
\newblock In {\em International Conference on Neural Information Processing
  Systems}, pages 155--162, 2005.

\bibitem{Bylinskii2018metric}
Z.~Bylinskii, T.~Judd, A.~Oliva, A.~Torralba, and F.~Durand.
\newblock What do different evaluation metrics tell us about saliency models?
\newblock {\em IEEE TPAMI}, 2018.

\bibitem{SAMresnet}
M.~Cornia, L.~Baraldi, G.~Serra, and R.~Cucchiara.
\newblock Predicting human eye fixations via an lstm-based saliency attentive
  model.
\newblock {\em IEEE TIP}, 2018.

\bibitem{erdem2013visual}
E.~Erdem and A.~Erdem.
\newblock Visual saliency estimation by nonlinearly integrating features using
  region covariances.
\newblock {\em Journal of vision}, 13(4):11--11, 2013.

\bibitem{LDS}
S.~Fang, J.~Li, Y.~Tian, T.~Huang, and X.~Chen.
\newblock Learning discriminative subspaces on random contrasts for image
  saliency analysis.
\newblock {\em IEEE TNNLS}, 28(5):1095--1108, 2017.

\bibitem{Garciadiaz2012On}
A.~Garciadiaz, V.~Lebor¨¢n, X.~R. Fdezvidal, and X.~M. Pardo.
\newblock On the relationship between optical variability, visual saliency, and
  eye fixations: a computational approach.
\newblock {\em Journal of Vision}, 12(7):17, 2012.

\bibitem{CAS}
S.~Goferman, L.~Zelnik-Manor, and A.~Tal.
\newblock Context-aware saliency detection.
\newblock {\em IEEE TPAMI}, 34(10):1915--1926, 2012.

\bibitem{harel2007graph}
J.~Harel, C.~Koch, and P.~Perona.
\newblock Graph-based visual saliency.
\newblock In {\em NIPS}, 2007.

\bibitem{hou2009dynamic}
X.~Hou and L.~Zhang.
\newblock Dynamic visual attention: Searching for coding length increments.
\newblock In {\em NIPS}, 2009.

\bibitem{huang2015salicon}
X.~Huang, C.~Shen, X.~Boix, and Q.~Zhao.
\newblock Salicon: Reducing the semantic gap in saliency prediction by adapting
  deep neural networks.
\newblock In {\em IEEE ICCV}, 2015.

\bibitem{itti1998model}
L.~Itti, C.~Koch, and E.~Niebur.
\newblock A model of saliency-based visual attention for rapid scene analysis.
\newblock {\em IEEE TPAMI}, 20(11):1254--1259, 1998.

\bibitem{jetley2016end}
S.~Jetley, N.~Murray, and E.~Vig.
\newblock End-to-end saliency mapping via probability distribution prediction.
\newblock In {\em IEEE CVPR}, 2016.

\bibitem{SALICON}
M.~Jiang, S.~Huang, J.~Duan, and Q.~Zhao.
\newblock Salicon: Saliency in context.
\newblock In {\em CVPR}, pages 1072--1080, 2015.

\bibitem{MIT300}
T.~Judd, F.~Durand, and A.~Torralba.
\newblock A benchmark of computational models of saliency to predict human
  fixations.
\newblock In {\em MIT Technical Report}, 2012.

\bibitem{MIT1003}
T.~Judd, K.~Ehinger, F.~Durand, and A.~Torralba.
\newblock Learning to predict where humans look.
\newblock In {\em IEEE ICCV}, 2010.

\bibitem{khaligh2014deep}
S.-M. Khaligh-Razavi and N.~Kriegeskorte.
\newblock Deep supervised, but not unsupervised, models may explain it cortical
  representation.
\newblock {\em PLoS computational biology}, 10(11):e1003915, 2014.

\bibitem{kienzle2007nonparametric}
W.~Kienzle, F.~A. Wichmann, M.~O. Franz, and B.~Sch{\"o}lkopf.
\newblock A nonparametric approach to bottom-up visual saliency.
\newblock In {\em NIPS}, 2007.

\bibitem{koch1987shifts}
C.~Koch and S.~Ullman.
\newblock Shifts in selective visual attention: towards the underlying neural
  circuitry.
\newblock In {\em Matters of intelligence}, pages 115--141. Springer, 1987.

\bibitem{kruthiventi2017deepfix}
S.~S. Kruthiventi, K.~Ayush, and R.~V. Babu.
\newblock Deepfix: A fully convolutional neural network for predicting human
  eye fixations.
\newblock {\em IEEE TIP}, 26(9):4446--4456, 2017.

\bibitem{kucuk2017large}
A.~Kucuk, J.~M. Banda, and R.~A. Angryk.
\newblock A large-scale solar dynamics observatory image dataset for computer
  vision applications.
\newblock {\em Scientific data}, 4:170096, 2017.

\bibitem{kummerer2014deep}
M.~K{\"u}mmerer, L.~Theis, and M.~Bethge.
\newblock Deep gaze i: Boosting saliency prediction with feature maps trained
  on imagenet.
\newblock {\em arXiv preprint arXiv:1411.1045}, 2014.

\bibitem{li2015visual}
G.~Li and Y.~Yu.
\newblock Visual saliency based on multiscale deep features.
\newblock In {\em IEEE CVPR}, pages 5455--5463, 2015.

\bibitem{SSD}
J.~Li, L.~Y. Duan, X.~Chen, T.~Huang, and Y.~Tian.
\newblock Finding the secret of image saliency in the frequency domain.
\newblock {\em IEEE TPAMI}, 37(12):2428--2440, 2015.

\bibitem{HFT}
J.~Li, M.~D. Levine, X.~An, X.~Xu, and H.~He.
\newblock Visual saliency based on scale-space analysis in the frequency
  domain.
\newblock {\em IEEE TPAMI}, 35(4):996--1010, 2013.

\bibitem{li2014visual}
J.~Li, Y.~Tian, and T.~Huang.
\newblock Visual saliency with statistical priors.
\newblock {\em IJCV}, 107(3):239--253, 2014.

\bibitem{li2015metric}
J.~Li, C.~Xia, Y.~Song, S.~Fang, and X.~Chen.
\newblock A data-driven metric for comprehensive evaluation of saliency models.
\newblock In {\em ICCV}, 2015.

\bibitem{PASCAL-S}
Y.~Li, X.~Hou, C.~Koch, J.~M. Rehg, and A.~L. Yuille.
\newblock The secrets of salient object segmentation.
\newblock In {\em IEEE CVPR}, 2014.

\bibitem{liu2018learning}
N.~Liu, J.~Han, T.~Liu, and X.~Li.
\newblock Learning to predict eye fixations via multiresolution convolutional
  neural networks.
\newblock {\em IEEE TNNLS}, 29(2):392--404, 2018.

\bibitem{pan2016shallow}
J.~Pan, E.~Sayrol, X.~Giro-i Nieto, K.~McGuinness, and N.~E. O'Connor.
\newblock Shallow and deep convolutional networks for saliency prediction.
\newblock In {\em IEEE CVPR}, 2016.

\bibitem{R2016Exploiting}
H.~R.-Tavakoli, A.~Borji, J.~Laaksonen, and E.~Rahtu.
\newblock Exploiting inter-image similarity and ensemble of extreme learners
  for fixation prediction using deep features.
\newblock {\em Neurocomputing}, 244, 2016.

\bibitem{riche2013saliency}
N.~Riche, M.~Duvinage, M.~Mancas, B.~Gosselin, and T.~Dutoit.
\newblock Saliency and human fixations: state-of-the-art and study of
  comparison metrics.
\newblock In {\em IEEE ICCV}, pages 1153--1160, 2013.

\bibitem{FES}
H.~R. Tavakoli, E.~Rahtu, and J.~Heikkil{\"a}.
\newblock Fast and efficient saliency detection using sparse sampling and
  kernel density estimation.
\newblock In {\em Scandinavian Conference on Image Analysis}, pages 666--675.
  Springer, 2011.

\bibitem{eDN}
E.~Vig, M.~Dorr, and D.~Cox.
\newblock Large-scale optimization of hierarchical features for saliency
  prediction in natural images.
\newblock In {\em IEEE CVPR}, 2014.

\bibitem{wang2018deep}
W.~Wang and J.~Shen.
\newblock Deep visual attention prediction.
\newblock {\em IEEE TIP}, 27(5):2368--2378, 2018.

\bibitem{yamins2014performance}
D.~L. Yamins, H.~Hong, C.~F. Cadieu, E.~A. Solomon, D.~Seibert, and J.~J.
  DiCarlo.
\newblock Performance-optimized hierarchical models predict neural responses in
  higher visual cortex.
\newblock {\em PNAS}, 111(23):8619--8624, 2014.

\bibitem{DUT-O}
C.~Yang, L.~Zhang, H.~Lu, R.~Xiang, and M.~H. Yang.
\newblock Saliency detection via graph-based manifold ranking.
\newblock In {\em IEEE CVPR}, 2013.

\bibitem{zhang2013saliency}
J.~Zhang and S.~Sclaroff.
\newblock Saliency detection: A boolean map approach.
\newblock In {\em IEEE ICCV}, pages 153--160, 2013.

\bibitem{zhang2008sun}
L.~Zhang, M.~H. Tong, T.~K. Marks, H.~Shan, and G.~W. Cottrell.
\newblock Sun: A bayesian framework for saliency using natural statistics.
\newblock {\em Journal of vision}, 8(7):32--32, 2008.

\bibitem{zhao2012learning}
Q.~Zhao and C.~Koch.
\newblock Learning visual saliency by combining feature maps in a nonlinear
  manner using adaboost.
\newblock {\em Journal of Vision}, 12(6):22--22, 2012.

\end{thebibliography}
}

\end{document}